\documentclass[conference]{IEEEtran}

\usepackage{cite}
\usepackage{amsmath,amssymb,amsfonts}
\usepackage{algorithm}
\usepackage{algorithmic}
\usepackage{graphicx}
\usepackage{textcomp}
\usepackage{xcolor}
\usepackage{booktabs}
\usepackage{multirow}
\usepackage{threeparttable}
\usepackage[caption=false,font=footnotesize]{subfig}

\def\BibTeX{{\rm B\kern-.05em{\sc i\kern-.025em b}\kern-.08em
    T\kern-.1667em\lower.7ex\hbox{E}\kern-.125emX}}

\begin{document}

\title{Joint Learning of Experiential Rules and Policies for Large Language Model Agents}

\author{
\IEEEauthorblockN{Shicheng Ye}
\IEEEauthorblockA{
\textit{Sun Yat-sen University} \\
yeschch@mail2.sysu.edu.cn}
\and
\IEEEauthorblockN{Chao Yu*}
\IEEEauthorblockA{
\textit{Sun Yat-sen University} \\
yuchao3@mail.sysu.edu.cn}
}

\maketitle
\pagestyle{plain}
\thispagestyle{plain}

\begin{abstract}
For LLM agents in multi-step interactive environments, a key challenge is to make effective use of accumulated interaction experience. Existing work has typically separated two uses of such experience: keeping it outside the model as natural-language rules for later prompting, or using trajectories and feedback to update the model parameters. The former is easy to interpret but can fall out of sync with the evolving policy; the latter improves the policy more broadly but provides only limited correction for local mistakes in sparse-reward settings. We present Joint Learning of Experiential Rules and Policies for LLM Agents (JERP), which updates a long-term experiential-rule pool and the policy from the same interaction trajectories. At decision time, JERP retrieves task-relevant rules and conditions the agent on them together with the interaction history. After each episode, it uses the collected trajectories both to optimize the policy and to revise the rule pool by comparing current rollouts with reference successful trajectories. This coupling keeps the rule pool aligned with the evolving policy while allowing stable and effective behaviors to be gradually absorbed into the model itself. Experiments on AlfWorld and WebShop show that JERP yields consistent gains in decision performance for complex interactive tasks.
\end{abstract}

\begin{IEEEkeywords}
large language model agents, learning from experience, reinforcement learning
\end{IEEEkeywords}

\section{Introduction}
\label{sec:co_insight-intro}

Large language models (LLMs) have become a natural foundation for general-purpose agents because they combine broad world knowledge with strong in-context learning and language reasoning abilities~\cite{plaat2025agentic}. In settings such as web navigation~\cite{ning2025survey}, embodied interaction~\cite{zeng2023large,liang2025large}, and tool use~\cite{xu2025llm}, however, solving a task rarely amounts to producing a single correct response. Agents often have to explore, recover from errors, and adjust their strategy over multiple steps before they succeed. In this setting, an important question is how an agent can accumulate experience during interaction and use it to make better decisions on future tasks~\cite{gao2025survey,du2026survey}.

Most existing approaches exploit interaction experience in one of two ways. One line of work converts past experience into reusable natural-language knowledge, including reflective texts~\cite{shinn2023reflexion}, guiding rules~\cite{zhao2024expel,fu2024autoguide}, and operation manuals~\cite{chen2024automanual}. In these methods, experience is externalized into a form that can be stored, retrieved, and revised, then fed back to the agent as additional context. The other line of work uses trajectories and environmental feedback to update model parameters directly, with the goal of improving the agent's policy on later tasks~\cite{du2026survey}.

These two uses of experience have complementary limitations. Retrieved rules can quickly address recurring local errors by reminding the agent of task preconditions or correcting persistent action biases~\cite{shinn2023reflexion,zhao2024expel,fu2024autoguide,chen2024automanual,liu2025contextual}. Yet rules do not execute themselves: they help only if the current model can interpret and apply them correctly in context. They can also become stale as the policy changes during training, and rules that were once useful may gradually lose value or even become misleading~\cite{xiong2025memory}. Parameter updates can improve behavior more globally, but feedback in complex interactive tasks is often sparse. As a result, some local but consequential mistakes may not be exposed early enough to receive timely and targeted correction~\cite{du2026survey,feng2025group}.

We therefore study whether the same interaction data can support both explicit rule updates and policy optimization. We propose Joint Learning of Experiential Rules and Policies for LLM Agents (JERP), which maintains a dynamically updated long-term experiential-rule pool and uses it together with the interaction history as decision context during each episode. Once an episode ends, the resulting trajectories are used twice: to optimize model parameters and to update the rule pool by comparing current behavior with reference successful trajectories. This allows the rule pool to evolve with the policy while gradually absorbing stable and repeatedly validated behaviors into the model itself.

\begin{figure*}[t]
  \centering
  \subfloat[Prompt optimization paradigm\label{fig:subfig-a}]
    {\includegraphics[width=0.285\linewidth]{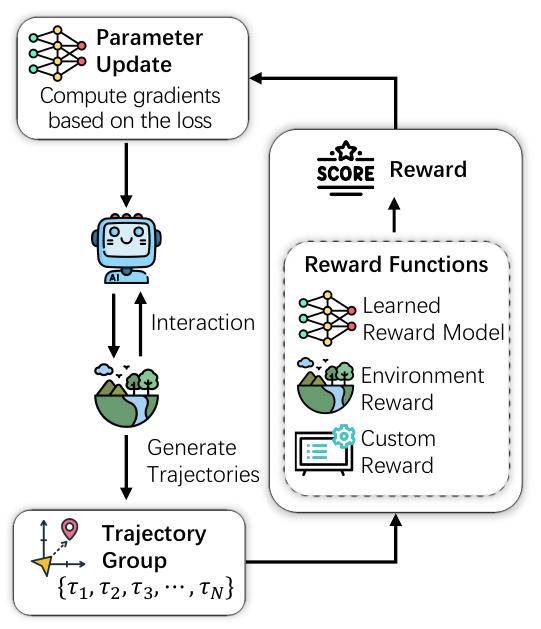}}
  \subfloat[Model-parameter optimization paradigm\label{fig:subfig-b}]
    {\includegraphics[width=0.275\linewidth]{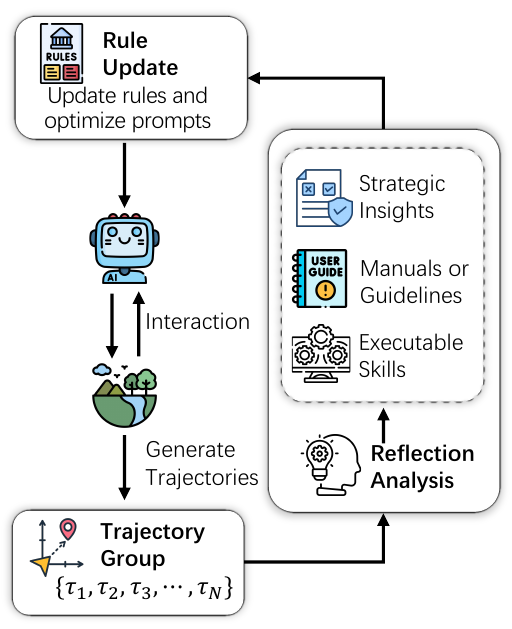}}
  \subfloat[JERP paradigm\label{fig:subfig-c}]
    {\includegraphics[width=0.42\linewidth]{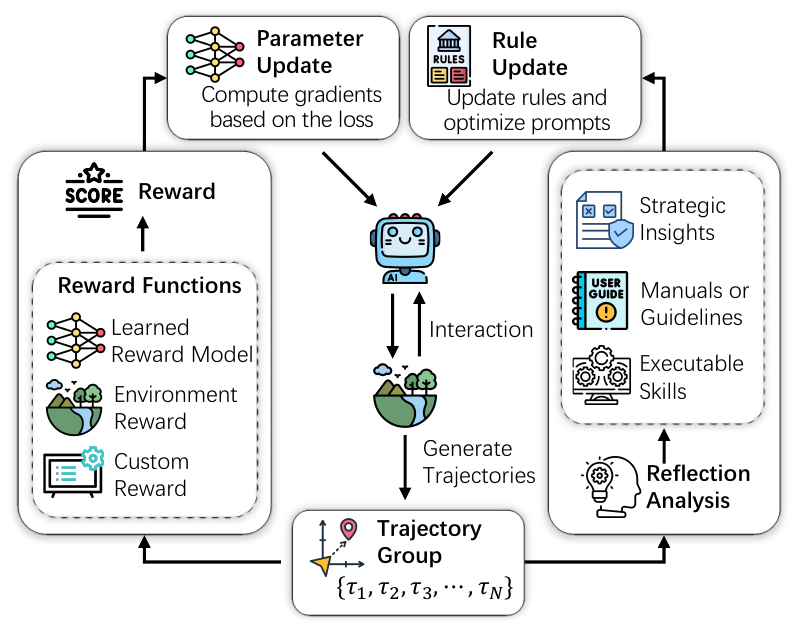}}
  \caption{Comparison between JERP and two basic paradigms.}
  \label{fig:framework-diff}
\end{figure*}

\section{Related Work}
\label{sec:co_insight-related}

\subsection{Experience Reuse through Prompt Engineering}

A common way to reuse interaction experience is to store it outside the model and feed it back as context in later episodes. Some methods preserve feedback or intermediate summaries after trial and error. For example, Reflexion~\cite{shinn2023reflexion} writes natural-language reflections after failures into episodic memory to guide future attempts. Generative Agents~\cite{park2023generative}, MemoryBank~\cite{zhong2024memorybank}, and MemGPT~\cite{packer2023memgpt} study how such memories can be retained, maintained, and retrieved over longer horizons.
Other studies distill interaction experience into reusable skills, rules, or manuals. For example, Voyager~\cite{wang2023voyager} accumulates experience and transfers it across tasks through an expanding library of code skills. ExpeL~\cite{zhao2024expel}, AutoGuide~\cite{fu2024autoguide}, and AutoManual~\cite{chen2024automanual} distill multi-task trial-and-error experience into reusable natural-language instructions for contextual guidance in subsequent tasks.

Beyond reflection-style memory, prior work has also induced tool-use patterns~\cite{schick2023toolformer}, workflows~\cite{wang2025agent}, and reusable skills~\cite{zheng2025skillweaver} from historical interactions, and has explored long-term use of experiential knowledge through knowledge bases~\cite{tang2025agent}, reflective memory~\cite{tan2025prospect}, and experience replay~\cite{liu2025contextual}. These studies establish that stored experience can improve later decisions when it is organized in a reusable form. Most of them, however, treat rule reuse and parameter learning as separate processes, leaving open how a rule pool should be maintained when the policy using it is also changing.

\subsection{Policy Optimization for LLMs via Reinforcement Learning Fine-Tuning}

Another line of work uses interaction experience to update the policy directly, turning trajectories, feedback, or preferences into optimization signals. Through reinforcement learning fine-tuning and closely related preference-optimization methods, these signals are used to update the model so that improved behavior is encoded in its parameters~\cite{du2026survey}. Early representative work such as InstructGPT~\cite{ouyang2022training} adopts a three-stage pipeline consisting of supervised fine-tuning, reward-model training, and PPO optimization, demonstrating that reinforcement learning from human feedback can substantially improve instruction following and output quality.

Subsequent studies have explored how to perform reinforcement learning fine-tuning for LLMs more stably and efficiently~\cite{tie2025survey}. In addition to PPO-based methods, GRPO~\cite{shao2024deepseekmath} estimates advantages by comparing samples within the same group, reducing the need for a separate value model and the associated memory overhead. DPO~\cite{rafailov2023direct} rewrites reinforcement learning from human feedback into a simpler direct preference-optimization objective. DMPO~\cite{shi2024direct} extends preference optimization to multi-turn agent tasks. For long-horizon interactive agent tasks, GiGPO~\cite{feng2025group} proposes a hierarchical group policy optimization method that combines episode-level and step-level relative advantage estimation to mitigate sparse rewards and fine-grained credit assignment. These methods improve the policy through parameter updates, but they do not maintain an explicit rule pool that can be inspected, revised, and reused during later interaction.

\section{Problem Definition and Basic Paradigms}
\label{sec:co_insight-problem}

We consider interactive tasks in which an LLM agent must choose actions from partial observations over multiple steps. Besides the policy parameters that produce actions, the agent also uses external experiential rules retrieved into the prompt. This section defines the resulting rule-augmented interaction process and introduces the two basic paradigms adopted by the two parts of JERP, namely prompt-based rule learning and group-relative policy optimization.

\subsection{Partially Observable Markov Decision Process}

Let $\mathcal{D}$ denote the task-instance space, with each instance $d \in \mathcal{D}$ drawn from distribution $p\left(d\right)$. For a fixed task instance, the agent interacts with the environment until the task terminates or the step budget is exhausted. Since the agent receives observations rather than full environment states, we model the interaction as a partially observable Markov decision process (POMDP), written as
\begin{equation}
  \mathcal{E} = \left(\mathcal{S}, \mathcal{A}, \mathcal{O}, P, \Omega, R, \gamma, T \right),
\end{equation}
where $\mathcal{S}$, $\mathcal{A}$, and $\mathcal{O}$ are the state, action, and observation spaces; $P\left(s_{t+1}\mid s_t,a_t\right)$ and $\Omega\left(o_{t+1}\mid s_{t+1},a_t\right)$ are the transition and observation functions; $R\left(s_t,a_t\right)$ is the reward function; $\gamma \in \left[0,1\right)$ is the discount factor; and $T$ is the maximum number of interaction steps. At step $t$, $s_t$ is the hidden environment state, $o_t$ is the observation available to the agent, and $a_t$ is the executed action.

Since the true state is not directly observed, the policy must condition on the interaction history rather than on $s_t$. We write the history at step $t$ as
\begin{equation}
  h_t = \left(o_0, a_0, o_1, \dots, a_{t-1}, o_t\right),
\end{equation}
where $o_0$ contains the initial task description.

JERP differs from a standard POMDP policy in that the prompt also contains reusable experience. For task instance \(d\), we maintain a long-term experiential-rule pool ${\mathcal{K}}\left(d\right)$, and denote the collection of such pools by $\mathcal{K}=\{{\mathcal{K}}\left(d\right)\}$. Before an episode starts, a size-controlled subset is selected from ${\mathcal{K}}\left(d\right)$ as the working rule set $\widetilde{\mathcal{K}}\left(d\right)$. Thus, the policy does not act from the history alone. Its decision context combines the task, the observed interaction history, and the currently retrieved rules.

The input prompt at step $t$ is
\begin{equation}
  x_t = \left(d, h_t, \widetilde{\mathcal{K}}\left(d\right)\right).
\end{equation}
The parameterized policy then samples an action by
\begin{equation}
\label{eq:policy}
  a_t \sim \pi_\theta\left(\cdot \mid x_t\right).
\end{equation}
After executing $a_t$, the environment returns reward $r_t = R\left(s_t,a_t\right)$, next observation $o_{t+1}$, and termination indicator $z_{t+1} \in \{0,1\}$. For a complete trajectory $\tau$, the discounted return is
\begin{equation}
  G\left(\tau\right) = \sum_{t=0}^{T-1}\gamma^t r_t.
\end{equation}

The learning problem is therefore not only to find better policy parameters. The rule pools also affect the trajectory distribution because they enter the prompt before each action. We define the training objective over both $\theta$ and $\mathcal{K}$ as
\begin{equation}
  J\left(\theta,\mathcal{K}\right)
  =
  \mathbb{E}_{d \sim p\left(d\right)}
  \Bigl[
  \mathbb{E}_{\tau \sim p_{\theta}\left(\tau \mid d,\mathcal{K}\right)}
  \bigl[
  G\left(\tau\right)
  \bigr]
  \Bigr],
\end{equation}
where $p_{\theta}\left(\tau \mid d,\mathcal{K}\right)$ is the trajectory distribution induced by $\pi_\theta$ when working rules are selected from the current rule pools. This notation makes the coupling explicit. Changing $\theta$ changes how the agent interprets the prompt, while changing $\mathcal{K}$ changes the prompt from which the same policy acts.

\subsection{Prompt-Based Experiential-Rule Learning}

One way to use experience is to keep it outside the model. Given trajectory samples $\left\{e_1,e_2,\dots,e_N\right\}$ from related tasks, prior prompt-based methods ask an LLM to summarize reusable behavioral guidance and store the result as natural-language rules. Abstractly,
\begin{equation}
\mathcal{K}=\operatorname{ExtractRules}\bigl(\{e_i\}_{i=1}^{N}\bigr).
\end{equation}

This paradigm is useful because the stored experience remains inspectable and editable, and can be revised without changing model parameters. Its limitation is equally important for our setting. The rule set is only effective through the current policy's ability to read and apply it. If the policy evolves during training while the rules remain static, the two may become poorly matched. This motivates treating rule maintenance as part of the training process rather than as a separate preprocessing step.

\subsection{Group-Relative Policy Optimization}

The other paradigm updates the policy from interaction feedback. For a given task instance $d$, group-relative methods sample $N$ complete trajectories $\left\{\tau_1, \tau_2, \dots, \tau_N\right\}$ under the old policy $\pi_{\theta_{\mathrm{old}}}$ and assign each trajectory a scalar reward $R\left(\tau_i\right)$. Instead of relying on an additional value model, they construct an advantage signal by comparing rewards within the sampled group.
\begin{equation}
A\left(\tau_i\right)=\operatorname{GroupComputation}\bigl(\{R\left(\tau_j\right)\}_{j=1}^{N}\bigr).
\end{equation}

For example, GRPO~\cite{shao2024deepseekmath} normalizes each trajectory reward by the mean and standard deviation of rewards in the same group. This style of optimization is suitable for our setting because it uses complete task outcomes and avoids training a separate value-function estimator. In JERP, the same sampled trajectory group is used not only for this policy update but also for updating the experiential-rule pool, which is the main departure from a pure parameter-optimization paradigm.

\section{Joint Learning of Experiential Rules and Policies}
\label{sec:co_insight-method}

We propose Joint Learning of Experiential Rules and Policies for LLM Agents (JERP), which updates policy parameters and the experiential-rule pool in the same training loop. In each episode, the current policy acts with a working rule set selected from the long-term rule pool. The sampled trajectories are then used for both a group-relative policy update and an experiential-rule-pool update. Fig.~\ref{fig:framework-diff} compares prompt-engineering-based experiential-rule learning, group-based reinforcement learning for policy optimization, and the overall structure of JERP.

\subsection{Overall Framework}

This work maintains experiential rules for each task. For a given task instance \(d\), its corresponding long-term experiential-rule pool is
\begin{equation}
  {\mathcal{K}}\left(d\right)=\left\{\left(z_i,s_i\right)\right\}_{i=1}^{|{\mathcal{K}}\left(d\right)|},
\end{equation}
where \(z_i\) is the textual content of the $i$-th experiential rule, and \(s_i\) is its utility score. The rule text records guidance information in natural language, while the utility score measures the rule's reference value for future decisions.

\begin{figure*}[t]
  \centering
  \includegraphics[width=1.0\linewidth]{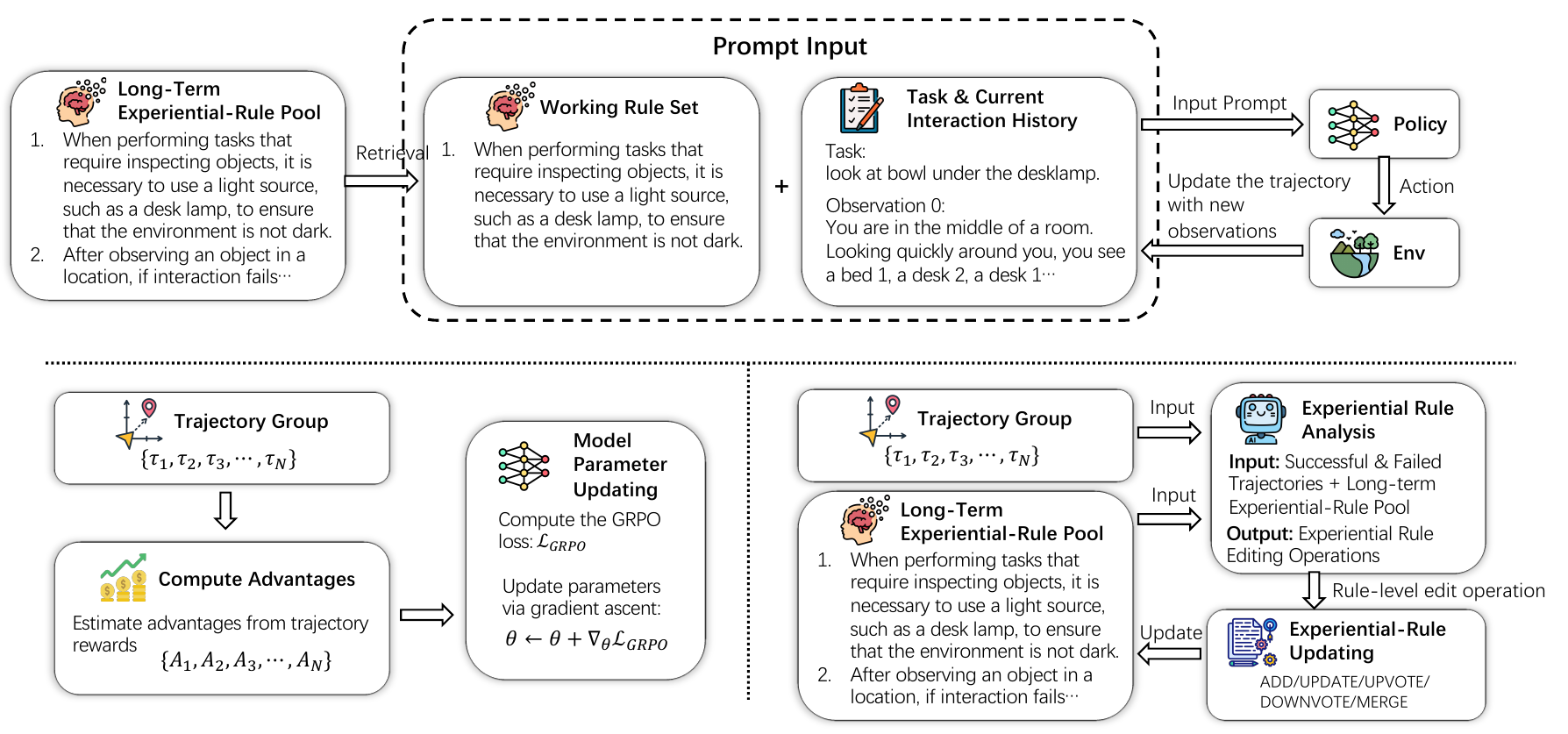}
  \caption{Overview of the JERP framework.}
  \label{fig:co-evolve-framework}
\end{figure*}

At the beginning of each episode, the system selects a subset from the current long-term experiential-rule pool \({\mathcal{K}}\left(d\right)\) as the working rule set $\widetilde{\mathcal{K}}\left(d\right)$, and provides it to the policy model together with the task description and current interaction history. The LLM agent then interacts with the environment on task instance \(d\) and generates a trajectory group \(\mathcal{T}_d\). The same trajectory group is used to update model parameters according to relative trajectory performance and to update the long-term experiential-rule pool by contrasting current trajectories with reference successful trajectories. The training process can be summarized as
\begin{equation}
  \mathcal{T}_d \sim \pi_\theta\left(\cdot \mid d,\widetilde{\mathcal{K}}\left(d\right)\right),
\end{equation}
\begin{align}
  \theta' &=
  \operatorname{Update}_{\theta}\left(\theta,\mathcal{T}_d\right), \\
  \mathcal{K}'\left(d\right) &=
  \operatorname{Update}_{\mathcal{K}}\bigl(
  \mathcal{K}\left(d\right),\mathcal{T}_d,\mathcal{T}^{+}\left(d\right)
  \bigr).
\end{align}
where parameter updates are driven by the trajectory group \(\mathcal{T}_d\), while rule updates combine \(\mathcal{T}_d\) with the reference successful trajectory set \(\mathcal{T}^{+}\left(d\right)\). The overall framework is shown in Fig.~\ref{fig:co-evolve-framework}.

\subsection{Trajectory Sampling}

Because the rule pool $\mathcal{K}\left(d\right)$ for task \(d\) expands during training, we do not feed all rules into the model in each episode. Instead, we select a size-controlled subset of rules for the current decision process. Since the working rules are already organized by task, the current implementation does not perform further instance-level relevance retrieval. The system sorts rules in \({\mathcal{K}}\left(d\right)\) by utility score in descending order and selects the top \(k\) rules as the input for the current episode,
\begin{equation}
  \widetilde{\mathcal{K}}\left(d\right)=\operatorname{topk}\bigl(\mathcal{K}\left(d\right),k\bigr).
\end{equation}
If the rule pool contains fewer than $k$ rules, all rules are used.

This score-based selection is an implementation choice; we do not claim it is generally superior to finer-grained instance-level retrieval. Within a single episode, the working rule set $\widetilde{\mathcal{K}}\left(d\right)$ remains fixed, avoiding context shifts caused by time-step-level rule reselection.

Rules in the working set are sorted by score and organized into the input prompt using a fixed template. Together with the task description and interaction history, they form the decision basis at the current time step. At time step $t$, the LLM policy defined in~\eqref{eq:policy} generates the action as
\begin{equation}
  a_t \sim \pi_{\theta}\left(\cdot \mid x_t\right),
\end{equation}
where $x_t=\left(d,h_t,\widetilde{\mathcal{K}}\left(d\right)\right)$ denotes the input prompt at the current time step.

\subsection{JERP Optimization Stage I: Model-Parameter Update}

For a fixed task instance $d$, we sample a trajectory group $\{\tau_i\}_{i=1}^{N}$ from the old policy $\pi_{\theta_{\mathrm{old}}}$. The $i$-th trajectory $\tau_i$ denotes a complete task-solving interaction episode. At time step $t$, its corresponding input prompt is
\begin{equation}
  x_{i,t} = \left(d,h_{i,t},\widetilde{\mathcal{K}}\left(d\right)\right).
\end{equation}
The agent generates action $a_{i,t}$ conditioned on this input and receives environment feedback until the trajectory terminates. In our setting, task rewards are returned only when a trajectory terminates. Therefore, for trajectory $\tau_i$, its trajectory-level reward is
\begin{equation}
R\left(\tau_i\right)=r^{(i)}_{|\tau_i|},
\end{equation}
where $r^{(i)}_{|\tau_i|}$ is the environment reward obtained by $\tau_i$ at termination.

For the same task instance $d$, we standardize the trajectory-level rewards $R\left(\tau_i\right)$ within the sampled group $\{\tau_i\}_{i=1}^{N}$ and construct group-relative advantages. Let the mean and standard deviation of trajectory rewards in the group be
\begin{equation}
\mu_d=\frac{1}{N}\sum_{j=1}^{N}R\left(\tau_j\right),
\qquad
\sigma_d=\sqrt{\frac{1}{N}\sum_{j=1}^{N}\bigl(R\left(\tau_j\right)-\mu_d\bigr)^2}.
\end{equation}
The group-relative advantage of the $i$-th trajectory is then defined as
\begin{equation}
A_i=\frac{R\left(\tau_i\right)-\mu_d}{\sigma_d+\delta},
\end{equation}
where $\delta$ is a numerical stability constant that prevents division by zero.

Following the advantage modeling strategy of GiGPO~\cite{feng2025group} for multi-step tasks, all time steps in the same trajectory share the same advantage,
\begin{equation}
A_{i,t}\equiv A_i.
\end{equation}

On this basis, the parameter-update objective is defined as
\begin{equation}
\mathcal{L}_{\mathrm{GRPO}}\left(\theta\right)
= -\frac{1}{N}\sum_{i=1}^{N}\frac{1}{|\tau_i|}
\sum_{t=0}^{|\tau_i|-1}
\ell_{i,t}\left(\theta\right),
\end{equation}
where the per-step surrogate term is
\begin{equation}
\begin{aligned}
\ell_{i,t}\left(\theta\right)
= {} & \min\!\Bigl(
\rho_{i,t}\left(\theta\right)A_{i,t},
\bar{\rho}_{i,t}\left(\theta\right)A_{i,t}
\Bigr) \\
& - \beta D_{\mathrm{KL}}\!\Bigl(
\pi_{\theta}\!\left(\cdot\mid x_{i,t}\right)
\,\|\,  
\pi_{\mathrm{ref}}\!\left(\cdot\mid x_{i,t}\right)
\Bigr),
\end{aligned}
\end{equation}
and
\begin{equation}
\bar{\rho}_{i,t}\left(\theta\right)=
\operatorname{clip}\!\bigl(
\rho_{i,t}\left(\theta\right),\,1-\epsilon,\,1+\epsilon
\bigr).
\end{equation}
with
\begin{equation}
\rho_{i,t}\left(\theta\right)=
\frac{
  \pi_{\theta}\!\left(a_{i,t}\mid x_{i,t}\right)
}{
  \pi_{\theta_{\mathrm{old}}}\!\left(a_{i,t}\mid x_{i,t}\right)
}.
\end{equation}
Here, $a_{i,t}$ denotes the action generated at time step $t$ of trajectory $\tau_i$; $x_{i,t}$ is the corresponding input prompt; $|\tau_i|$ is the number of environment interaction steps in $\tau_i$; $\epsilon$ is the clipping coefficient; $\beta$ is the coefficient of KL regularization; and $\pi_{\mathrm{ref}}$ denotes the reference policy.

This stage updates model parameters according to the relative performance of trajectories under the given task instance and working rule set.

\subsection{JERP Optimization Stage II: Experiential-Rule Updating via Contrastive Reflection}

In the experiential-rule updating stage, the long-term experiential-rule pool is maintained rather than only expanded. Given the current long-term experiential-rule pool, the trajectory group sampled in the current episode, and reference successful trajectories for the same task, the update needs to extract reusable experience, revise inaccurate or weak rules, merge semantically similar and functionally overlapping rules, and keep the rule pool within its capacity limit.

Because the content of the rule pool is represented in natural language, rule addition and revision are difficult to formulate as a standard numerical optimization problem. We adopt a structured update method based on contrastive reflection. The system generates rule-level edit operations from the current rule pool, current trajectory group, and reference successful trajectories, and then obtains the updated long-term experiential-rule pool. This idea is close to ExpeL~\cite{zhao2024expel}, which maintains natural-language rules through experiential reflection; both approaches prompt an LLM to generate rule-update operations. We further introduce a merge operation to combine semantically similar and functionally overlapping rules when the long-term experiential-rule pool expands during online training.

For task instance $d$, let its long-term experiential-rule pool be $\mathcal{K}\left(d\right)$ and let the trajectory group sampled in the current episode be
\begin{equation}
\mathcal{T}_d=\{\tau_i\}_{i=1}^{N}.
\end{equation}
The reference successful trajectory set for the same task instance is
\begin{equation}
  \mathcal{T}^{+}\left(d\right)=\{\tau_1^{+},\tau_2^{+},\dots,\tau_M^{+}\},
\end{equation}
where $\mathcal{T}^{+}\left(d\right)$ denotes the successful trajectories accumulated for the same task during training, and $M=\left|\mathcal{T}^{+}\left(d\right)\right|$ denotes the number of reference successful trajectories for task instance $d$. We do not use human demonstrations or successful trajectories from the test stage. If no successful trajectory has been accumulated for the current task instance, we set $\mathcal{T}^{+}\left(d\right)=\varnothing$; in this case, rule-pool updating relies only on the current rule pool and current trajectory group.

Based on these inputs, the system generates a set of rule-level edit operations,
\begin{equation}
  \mathcal{O}\left(d\right)
  =
  \operatorname{ReflectAndEdit}\bigl(
  \mathcal{K}\left(d\right),
  \mathcal{T}_d,
  \mathcal{T}^{+}\left(d\right)
  \bigr),
\end{equation}
where $\operatorname{ReflectAndEdit}(\cdot)$ denotes the rule-update operator based on contrastive reflection. In our implementation, this operator prompts an LLM with frozen parameters to generate a structured list of rule-edit operations from the current rule pool, current trajectory group, and reference successful trajectories. The output is constrained to a fixed edit format so that it can be parsed and applied automatically.

\begin{algorithm}[t]
\caption{JERP Training Procedure}
\label{alg:co_rule}
\textbf{Input:} Training task set $\mathcal{D}$, initial policy parameters $\theta$, working rule set size $k$, trajectory group size $N$, maximum interaction steps $T$, training steps $E$.
\begin{algorithmic}[1]
\FOR{each task $d \in \mathcal{D}$}
    \STATE Initialize long-term experiential-rule pool $\mathcal{K}(d) \leftarrow \varnothing$
    \STATE Initialize reference successful trajectory set $\mathcal{T}^{+}(d) \leftarrow \varnothing$
\ENDFOR
\FOR{$e = 1,2,\dots,E$}
    \STATE Sample task instance $d \in \mathcal{D}$
    \STATE Set $\theta_{\mathrm{old}} \leftarrow \theta$
    \STATE Construct working rule set $\widetilde{\mathcal{K}}(d) \leftarrow \operatorname{topk}(\mathcal{K}(d), k)$
    \STATE Initialize trajectory group $\mathcal{T}_d \leftarrow \varnothing$
    \FOR{$i = 1,2,\dots,N$}
        \STATE Initialize $\tau_i \leftarrow \varnothing$ and $h_{i,0} \leftarrow \{o_{i,0}\}$
        \FOR{$t = 0,1,\dots,T-1$}
            \STATE Set $x_{i,t} \leftarrow (d,h_{i,t},\widetilde{\mathcal{K}}(d))$
            \STATE Sample action $a_{i,t} \sim \pi_{\theta_{\mathrm{old}}}(\cdot \mid d, h_{i,t}, \widetilde{\mathcal{K}}(d))$
            \STATE Execute $a_{i,t}$ and update $h_{i,t+1}$
            \STATE $\tau_i \leftarrow \tau_i \cup \{(x_{i,t},a_{i,t},h_{i,t+1})\}$
            \IF{$\tau_i$ terminates}
                \STATE \textbf{break}
            \ENDIF
        \ENDFOR
        \STATE Compute trajectory reward $R(\tau_i)$
        \STATE $\mathcal{T}_d \leftarrow \mathcal{T}_d \cup \{\tau_i\}$
    \ENDFOR
    \STATE Compute group-relative advantages from trajectory rewards
    \STATE Update $\theta$ using the GRPO objective
    \STATE $\mathcal{O}(d) \leftarrow \operatorname{ReflectAndEdit}(\mathcal{K}(d), \mathcal{T}_d, \mathcal{T}^{+}(d))$
    \STATE Apply rule updates $\mathcal{K}(d) \leftarrow \operatorname{Apply}(\mathcal{K}(d), \mathcal{O}(d))$
    \STATE Update rule scores and prune $\mathcal{K}(d)$ under the capacity limit
    \STATE $\mathcal{T}^{+}(d) \leftarrow \mathcal{T}^{+}(d) \cup \{\tau_i \in \mathcal{T}_d \mid R(\tau_i)>0\}$
\ENDFOR
\STATE \textbf{return} $\theta$ and $\{\mathcal{K}(d)\}_{d\in\mathcal{D}}$
\end{algorithmic}
\end{algorithm}

To balance result parseability and online maintenance cost, we use a small and fixed structured edit syntax to represent rule-pool updates. The system generates five types of primitive operations, \texttt{ADD}$(z)$, \texttt{EDIT}$(q,z)$, \texttt{UPVOTE}$(q)$, \texttt{DOWNVOTE}$(q)$, and \texttt{MERGE}$(Q, z)$. Here, $q$ denotes the identifier of a target rule, $Q$ denotes a set of rule identifiers to be merged, and $z$ denotes newly generated rule text. In \texttt{ADD} and \texttt{EDIT}, $z$ corresponds to the newly added rule text and the revised rule text, respectively. In \texttt{MERGE}, $z$ corresponds to the merged rule text.

These five operations correspond to supplementation, revision, reinforcement, weakening, and compression in rule-pool maintenance. Specifically, \texttt{ADD} introduces supplementary experience, \texttt{EDIT} modifies existing rules, \texttt{UPVOTE} and \texttt{DOWNVOTE} increase or decrease the utility score of a target rule, and \texttt{MERGE} combines semantically close and functionally overlapping rules into one rule.

After generating structured edit operations, the system first edits rule texts according to the operation list generated in the current episode,
\begin{equation}
  {\mathcal{K}}\left(d\right) \leftarrow \operatorname{Apply}\bigl(\mathcal{K}\left(d\right),\mathcal{O}\left(d\right)\bigr).
\end{equation}
The system then updates the utility score of each rule to reflect its future retention value. We adopt a fixed-step update rule. For \texttt{ADD}, the new rule is assigned an initial score $s_0>0$. For \texttt{UPVOTE} and \texttt{DOWNVOTE}, the target rule score is adjusted by fixed increments $\delta_{+}$ and $\delta_{-}$, respectively. For \texttt{EDIT}, the edited rule retains its original score. For \texttt{MERGE}, the score of the merged rule is set to the maximum score among the original rules. If a rule's updated score is below a predefined threshold, it is removed from the rule pool.

Through these edit and score updates, the long-term experiential-rule pool incorporates useful experience from the current trajectory group while removing or merging rules with lower retention value.

Algorithm~\ref{alg:co_rule} summarizes the full training process, including trajectory sampling, parameter updating, and rule-pool updating.

\section{Experiments}

This section presents the experimental evaluation on two interactive benchmarks and analyzes where the gains of JERP come from. We first compare JERP with baselines that use online reasoning, reflection-style memory, or policy optimization. We then examine the effect of continual experiential-rule-pool updating through an ablation experiment.

\subsection{Environments and Evaluation Metrics}

We evaluate JERP on two interactive environments, AlfWorld~\cite{shridhar2020alfworld} and WebShop~\cite{yao2022webshop}.
AlfWorld is a text-based household manipulation environment. We use its six task types, \textit{pick and place}, \textit{pick clean then place}, \textit{pick heat then place}, \textit{pick cool then place}, \textit{look at obj}, and \textit{pick two obj}, abbreviated as \textit{Pick}, \textit{Clean}, \textit{Heat}, \textit{Cool}, \textit{Look}, and \textit{Pick2}. The environment is partially observable and requires multi-step interaction.
WebShop is an interactive environment for online shopping tasks. Given a user's product requirements, the agent performs search, browsing, comparison, and purchase actions through Web-page interactions.
We use environment-specific evaluation metrics. For AlfWorld, we report task success rate (\%). For WebShop, we report average score and task success rate (\%).

\subsection{Baseline Methods}

We compare JERP with five baselines, Vanilla LLM, ReAct~\cite{yao2022react}, Reflexion~\cite{shinn2023reflexion}, RLOO~\cite{ahmadian2024back}, and GRPO~\cite{shao2024deepseekmath}, covering direct prompting, online reasoning, reflection-style memory reuse, and group-relative policy optimization.
Vanilla LLM directly prompts the base model to act without reasoning scaffolding, cross-episode memory, or parameter updates.
ReAct alternates between reasoning and action during interaction. It relies on observations and context from the current episode, without cross-episode memory or parameter updates.
Reflexion writes natural-language reflections from prior attempts into memory to guide later decisions. Compared with ReAct, it reuses cross-episode experience, but this experience remains in external memory rather than in model parameters.
RLOO updates policy parameters with a REINFORCE-style objective, using a leave-one-out baseline computed from the other trajectories in the same sampled group to estimate the advantage.
GRPO updates policy parameters through group-relative trajectory comparison without relying on an independent value model. Unlike ReAct and Reflexion, RLOO and GRPO improve later decisions through training rather than through test-time reasoning or memory alone.

\begin{table}[htbp]
\centering
\caption{Hyperparameter configurations of JERP in different benchmark environments.}
\label{tab:jerp-hyperparameters}
\begin{tabular*}{\columnwidth}{@{\extracolsep{\fill}}lcc@{}}
\toprule
Hyperparameter & AlfWorld & WebShop \\
\midrule
Learning rate & 3e-6 & 3e-6 \\
Group sampling size & 8 & 8 \\
Reward discount factor & 0.95 & 0.95 \\
Training epochs & 200 & 300 \\
Batch size & 256 & 64 \\
LoRA rank & 64 & 64 \\
LoRA scaling factor & 64 & 64 \\
Maximum steps per episode & 50 & 15 \\
\bottomrule
\end{tabular*}
\end{table}

\begin{table*}[!t]
\centering
\begin{threeparttable}[c]
\caption{Performance comparison between JERP and baseline methods on two benchmark environments.}
\small
\setlength{\tabcolsep}{4.5pt}
\renewcommand{\arraystretch}{1.1}
\begin{tabular}{lccccccc|cc}
\toprule
\multirow{2}{*}{Method}
& \multicolumn{7}{c|}{AlfWorld\tnote{a}}
& \multicolumn{2}{c}{WebShop\tnote{b}} \\
          & Pick & Look & Clean & Heat & Cool & Pick2 & All & Score & Success Rate \\
\midrule
Vanilla LLM & 5.9 & 5.5 & 3.3 & 9.7 & 4.2 & 0.0 & 4.1 & 23.1 & 5.2 \\
ReAct      & 17.4 & 20.5 & 15.7 & 6.2 & 7.7 & 2.0 & 12.8 & 40.1 & 11.3 \\
Reflexion  & 35.3 & 22.2 & 21.7 & 13.6 & 19.4 & 3.7 & 21.8 & 55.8 & 21.9 \\
RLOO(+LoRA) & 71.5 & 68.3 & 61.2 & 34.4 & 41.0 & 19.9 & 48.7 & 71.9 & 57.8 \\
GRPO(+LoRA) & 78.5 & 73.3 & 50.7 & 62.7 & 51.7 & 33.9 & 57.8 & 78.1 & 56.2 \\
JERP(+LoRA) & 72.2 & 69.8 & 65.4 & 67.4 & 60.1 & 42.5 & 61.5 & 79.0 & 64.1 \\
\bottomrule
\end{tabular}
\begin{tablenotes}
  \item [a] AlfWorld uses task success rate (\%) as the evaluation metric.
  \item [b] WebShop uses average score and task success rate (\%) as evaluation metrics.
\end{tablenotes}
\label{tab:main_result}
\end{threeparttable}
\end{table*}

\subsection{Implementation Details}

All experiments are conducted on a server running Ubuntu 22.04 LTS with two Intel Xeon Gold 6226R processors, approximately 512 GB of memory, and four NVIDIA A30 GPUs. For GRPO and JERP, which require reinforcement learning training, we use LoRA~\cite{hu2022lora} for parameter-efficient fine-tuning. Unless otherwise specified, all reported results are averaged over three independent runs on the same platform. The key hyperparameters are shown in Table~\ref{tab:jerp-hyperparameters}.

\subsection{Comparison with Baselines}

\begin{figure*}[t]
  \centering
  \includegraphics[width=1.0\textwidth]{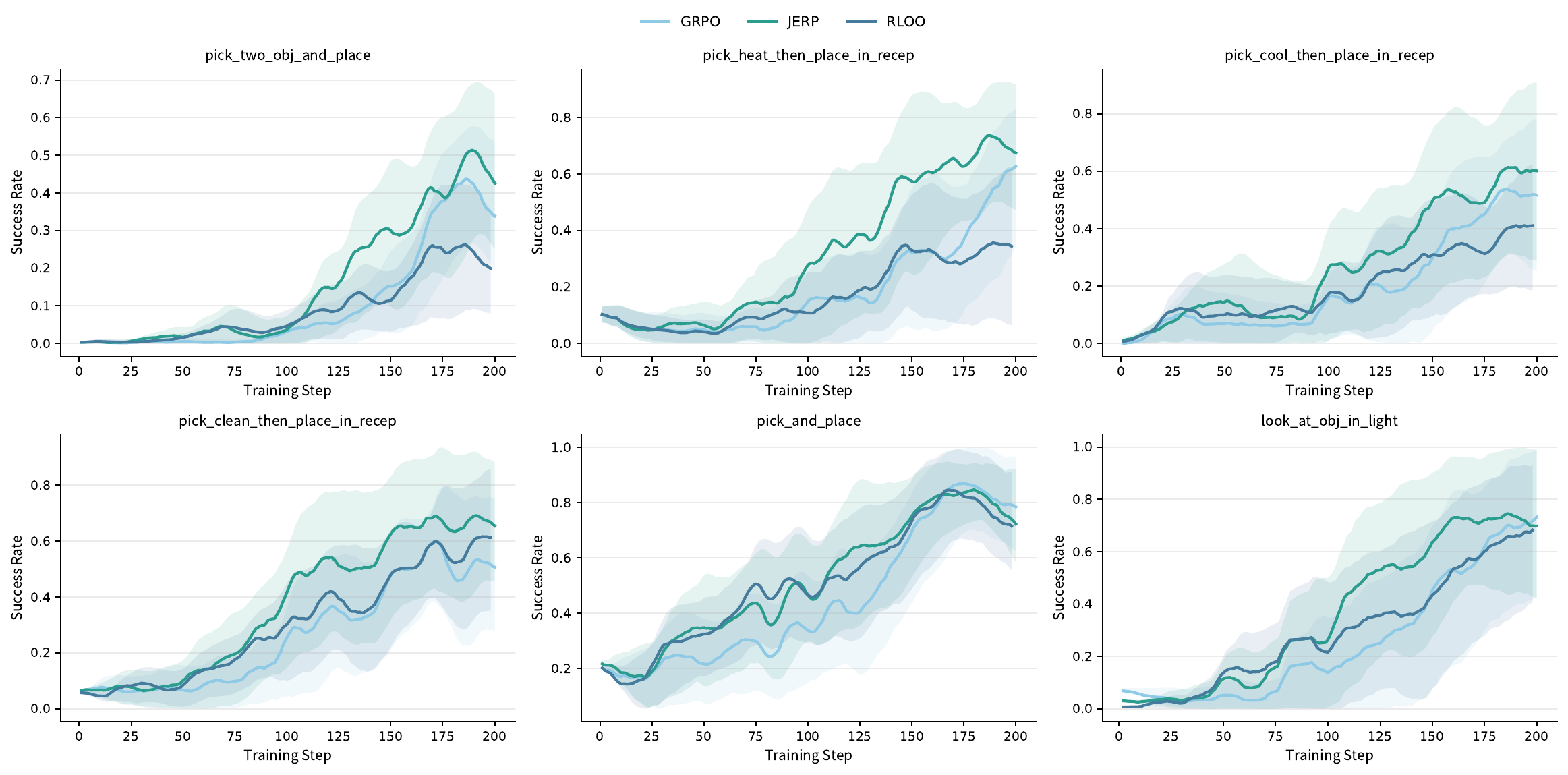}
  \caption{Success-rate curves of JERP and baseline methods on different AlfWorld task types over training steps.}
  \label{fig:jerp-alfworld}
\end{figure*}

Table~\ref{tab:main_result} and Fig.~\ref{fig:jerp-alfworld} show the performance differences between JERP and the baselines. Vanilla LLM performs poorly on both environments (4.1\% success rate on AlfWorld and 5.2\% on WebShop), confirming that direct prompting of the base model is insufficient for these multi-step tasks and that the gains of the other methods come from how they use interaction experience. On AlfWorld, JERP obtains the best overall success rate at 61.5\%, above GRPO at 57.8\% and RLOO at 48.7\%. The gains over GRPO are concentrated on \textit{Clean}, \textit{Heat}, \textit{Cool}, and \textit{Pick2}, while GRPO remains higher on \textit{Pick} and \textit{Look}. On WebShop, JERP also performs best, with the highest average score (79.0) and a success rate of 64.1\%, clearly above both RLOO (57.8\%) and GRPO (56.2\%).

The comparison also separates the roles of different forms of experience use. ReAct mainly relies on context within the current episode and does not reuse cross-episode experience. Reflexion adds language-based memory, but the stored reflections remain outside parameter learning. RLOO and GRPO improve the policy through reinforcement learning and are much stronger than ReAct and Reflexion, but they use trajectories only for parameter updates. JERP adds a second use of the same trajectories by maintaining the experiential-rule pool during training, which is consistent with its stronger results on the more constraint-heavy AlfWorld categories and on WebShop success rate.

The training curves in Fig.~\ref{fig:jerp-alfworld} show a similar pattern. JERP and GRPO are close at the early stage, but JERP improves more clearly in the middle and late stages on \textit{Heat}, \textit{Cool}, \textit{Clean}, and \textit{Pick2}. These tasks usually require longer operation sequences and more intermediate constraints, where reusable rules can preserve effective operation patterns observed in earlier episodes.

These results suggest that rule-pool updating is most helpful when the task exposes recurring intermediate mistakes. Parameter updates still provide the main policy improvement, while the rule pool keeps task-specific corrections available during later trajectory sampling.

\subsection{Ablation on the Key Mechanism}

To analyze the role of dynamic experiential-rule-pool updating, we construct an ablated variant that pauses rule-pool updates. This variant uses the same training pipeline as JERP, but updates the rule pool only once at the initial training stage. The rule-pool content is then fixed and no longer supplemented or revised according to later interaction trajectories.

\begin{figure}[H] 
  \centering
  \includegraphics[width=0.8\linewidth]{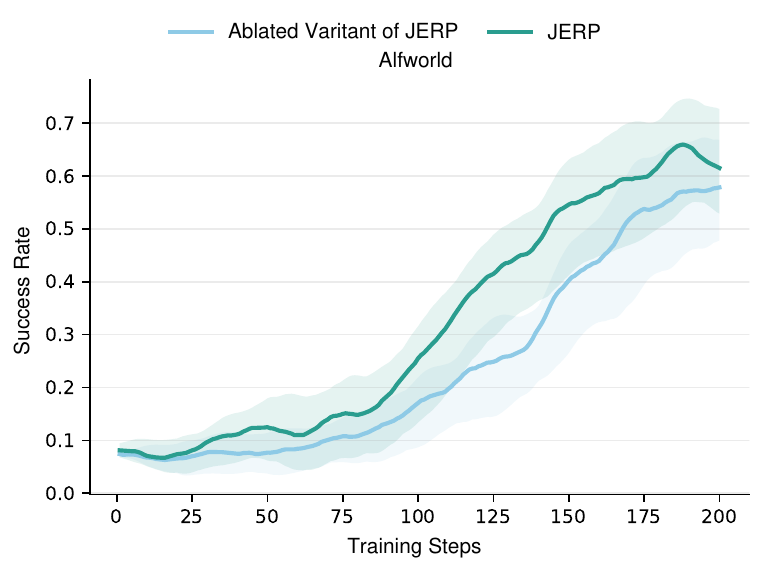}
  \caption{Success-rate curves of JERP and its ablated variant on AlfWorld over training steps.}
  \label{fig:jerp-alfworld-melt}
\end{figure}

The results are shown in Fig.~\ref{fig:jerp-alfworld-melt}. The complete JERP maintains a higher task success rate throughout training. The ablated variant still benefits from rules formed at the initial stage, but its improvement slows after the rule pool is fixed. As the policy changes, later trajectories expose different errors and useful behavior patterns. A fixed rule pool cannot absorb these later observations, whereas JERP continues to revise the rule pool during training.

\section{Conclusion}

This work proposes JERP, a joint learning method for experiential rules and policies in LLM agents.
The method maintains a long-term experiential-rule pool together with a working rule set for each task. During training, the same batch of interaction trajectories is used both for group-relative policy optimization and for experiential-rule updating. This design keeps rule maintenance and parameter learning in the same loop, so that reusable local experience can be preserved as rules while better behavior is gradually absorbed into the policy.
Experiments on AlfWorld and WebShop show that JERP improves overall performance over the baselines. The gains are more evident on tasks with longer decision sequences and richer intermediate constraints. Ablation results further show that continual experiential-rule-pool updating is important to these improvements.
Future work may explore more adaptive rule-retrieval strategies and extend the method to multi-agent interactive settings.

\bibliographystyle{IEEEtran}
\bibliography{refs}

\end{document}